\documentclass[letterpaper, 10 pt, conference]{ieeeconf}  

\usepackage{graphicx} 
\usepackage{multirow}
\usepackage[table,xcdraw]{xcolor}
\usepackage{amsmath}
\usepackage{amssymb}
\usepackage{bm}

\IEEEoverridecommandlockouts                       

\title{\LARGE \bf
Self-Aligning Depth-regularized Radiance Fields for Asynchronous RGB-D Sequences
}

\author{Yuxin Huang$^{1, 2}$, Andong Yang$^{3}$, Zirui Wu$^{1, 2}$, Yuantao Chen$^{1, 4}$, Runyi Yang$^{1, 5}$,\\ \qquad\quad Zhenxin Zhu$^{1, 6}$, Chao Hou$^{1, 7}$, Hao Zhao$^{1}$, and Guyue Zhou$^{1}$
\thanks{This work was sponsored by Tsinghua-Toyota Joint Research Fund (20223930097).}
\thanks{$^{1}$Hao Zhao and Guyue Zhou are with AIR, Tsinghua University, Beijing, China \texttt{\{zhaohao, zhouguyue\}@air.tsinghua.edu.cn}.}
\thanks{$^{2}$Yuxin Huang and Zirui Wu are with Beijing Institute of Technology, Beijing, China \texttt{\{yuxinhuang, wuzirui\}@bit.edu.cn}.}
\thanks{$^{3}$Andong Yang is with the Research Center for Intelligent Computing Systems, Institute of Computing Technology, Chinese Academy of Sciences, University of Chinese Academy of Sciences, Beijing, China \texttt{yangandong19b@ict.ac.cn}.}
\thanks{$^{4}$Yuantao Chen is with Xi’an University of Architecture and Technology, Xi’an, China \texttt{yuantao@xauat.edu.cn}.}
\thanks{$^{5}$Runyi Yang is with Imperial College London, London, England \texttt{runyi.yang23@imperial.ac.uk}.}
\thanks{$^{6}$Zhenxin Zhu is with Beihang University, Beijing, China \texttt{zhuzhenxin@buaa.edu.cn
}}
\thanks{$^{7}$Chao Hou is with The University of Hong Kong, Hong Kong SAR \texttt{houchao@connect.hku.hk}}
\thanks{Correspondence: Hao Zhao, Guyue Zhou.}
}

\begin{document}
\maketitle
\thispagestyle{empty}
\pagestyle{empty}

\begin{abstract}
It has been shown that learning radiance fields with depth rendering and depth supervision can effectively promote the quality and convergence of view synthesis. However, this paradigm requires input RGB-D sequences to be synchronized, hindering its usage in the UAV city modeling scenario. As there exists asynchrony between RGB images and depth images due to high-speed flight, we propose a novel time-pose function, which is an implicit network that maps timestamps to $\rm SE(3)$ elements. To simplify the training process, we also design a joint optimization scheme to jointly learn the large-scale depth-regularized radiance fields and the time-pose function. Our algorithm consists of three steps: (1) time-pose function fitting, (2) radiance field bootstrapping, (3) joint pose error compensation and radiance field refinement. In addition, we propose a large synthetic dataset with diverse controlled mismatches and ground truth to evaluate this new problem setting systematically. Through extensive experiments, we demonstrate that our method outperforms baselines without regularization. We also show qualitatively improved results on a real-world asynchronous RGB-D sequence captured by drone. Codes, data, and models will be made publicly available.
\end{abstract}

\section{Introduction}
\label{sec:intro}

Incorporating depth rendering and depth supervision into radiance fields has been demonstrated as a helpful regularization technique in several recent studies \cite{deng2022deep, roessle2022dense, xiangli2022bungeenerf, rematas2022urban, wu2023mars}. However, this technique has not yet been successfully introduced into radiance field learning from UAV (Unmanned Aerial Vehicle) images, despite its significance in many robotics applications. A closer look at the aforementioned works indicates that they presume synchronized RGB and depth signals, which is hard to guarantee in UAV vision due to potential temporal asynchrony caused by high-speed flight, leading to slight discrepancies between RGB images and depth images. Therefore, we study the \emph{problem} of learning depth-regularized radiance fields from asynchronous RGB-D sequences.

As a recap, the canonical radiance field \cite{mildenhall_nerf_2020} learns a neural network parameterized by \bm{$\theta$} that represents a 3D scene from input images $I$ and their intrinsic/extrinsic parameters $\mathcal{T}_{\rm I}$. To alleviate the reliance on $\mathcal{T}_{\rm I}$, some works \cite{wang2021nerf, lin2021barf, jeong2021self} aim to resolve a different problem that self-calibrates $\mathcal{T}_{\rm I}$. In other words, they jointly learn \bm{$\theta$} and $\mathcal{T}_{\rm I}$ from input images $I$. Similarly, the \emph{formulation} considered here is to learn scene representation \bm{$\theta$}, camera parameters $\mathcal{T}_{\rm I}$ and $\mathcal{T}_{\rm D}$ from inputs RGB images $I$ and depth images $D$. 

\begin{figure}[!t]
\centering
\includegraphics[width=0.47\textwidth]{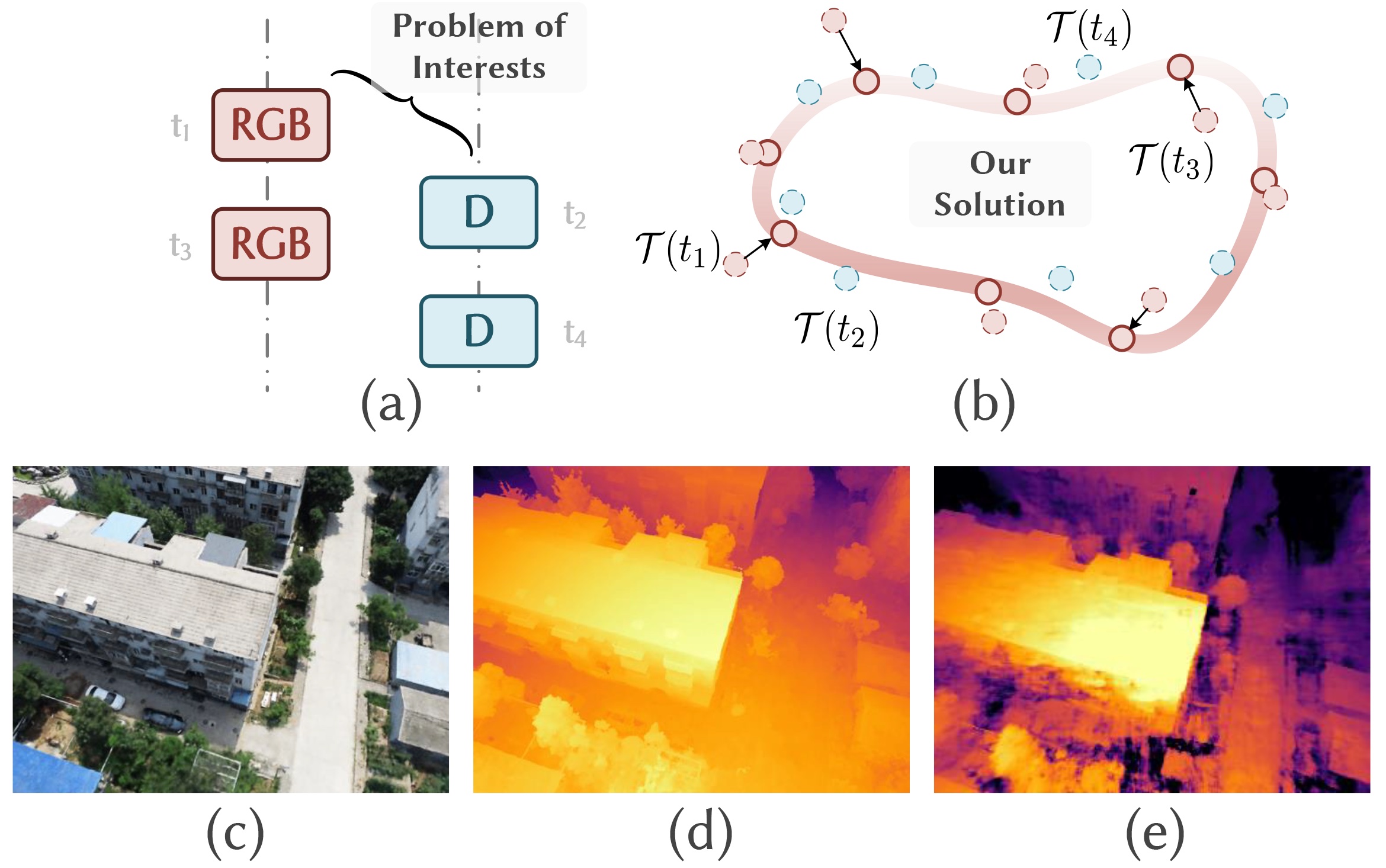} 
\caption{To learn a depth-regularized radiance field using (a) asynchronous RGB-D sequences, we propose a (b) time-pose function to map from timestamp to camera pose. For a (c) novel view, our method can render a better (d) depth map than (e) Mega-NeRF.}
\label{fig:teaser}
\end{figure}

We leverage an important prior specific to this problem: RGB-D frames are actually sampled from the same physical trajectory. As conceptually shown in Fig. \ref{fig:teaser}-a/b, $\mathcal{T}_{\rm I}$ and $\mathcal{T}_{\rm D}$ can be considered as samples from a function that maps timestamps to $\rm SE(3)$ elements. We name this function as \textbf{time-pose function} and model it with a neural network parameterized by \bm{$\phi$}. As such, we address the \emph{problem} with a \emph{new formulation} that learns scene representation \bm{$\theta$} and the time-pose function \bm{$\phi$} from inputs RGB images $I$ and depths $D$. An interesting fact is that both \bm{$\theta$} and \bm{$\phi$} are implicit neural representation networks (or coordinate-based networks) that allow fully differentiable training. To our knowledge, this \emph{new formulation} has not been proposed before.

We also propose an effective learning scheme. In the \textbf{first} stage, we fit the time-pose function \bm{$\phi$} using one modality (e.g., RGB images) and infer the poses of the other using a balanced pose regression loss and a speed regularization term. \textbf{Secondly}, we bootstrap a large-scale radiance field \bm{$\theta$} based upon Mega-NeRF \cite{turki_mega-nerf_2022} using the outputs of the trained time-pose function. Thanks to the time-pose function, depth regularization can be imposed here in spite of RGB-D misalignment. \textbf{Thirdly}, thanks to the cascade of two fully differentiable implicit representation networks, we jointly optimize the 3D scene representation \bm{$\theta$} and compensate for pose errors by updating \bm{$\phi$}. 

Since the \emph{problem} considered is new, we contribute a synthetic dataset (named AUS) for systematic evaluation. Using six large-scale 3D scenes, realistic drone trajectories of different difficulty levels are generated. Specifically, simple trajectories are heuristically designed with a zig-zag pattern, while complicated ones are generated by manual control signals in simulation. We also control the mismatch between RGB-D sequences using different protocols to cover as many scenarios as possible that the algorithm may encounter in reality. Through a set of comprehensive experiments, we show the proposed method outperforms baseline methods without regularization and our design choices contribute positively to performance. Last but not least, we present a real-world evaluation using asynchronous sensors on the drone. Our depth rendering results (on unseen viewpoint) are shown in Fig.~\ref{fig:teaser}-d, which is much better than the result of Mega-NeRF shown in Fig.~\ref{fig:teaser}-e. This success is credited to the usage of depth regularization, which was made possible by our novel algorithm.

To summarize, we have the following contributions in this paper: (1) We formalize the new \emph{problem} of learning depth-regularized radiance fields from asynchronous RGB-D sequences, which is rooted in many UAV applications. (2) We identify an essential domain-specific prior in this problem: RGB-D frames are sampled from the same underlying trajectory. We instantiate this prior into a novel time-pose function and develop a cascaded, fully differentiable implicit representation network. (3) In order to systematically evaluate the task, we contribute a photo-realistically rendered synthetic dataset that simulates different types of mismatch. (4) Through benchmarking on this new dataset and real-world asynchronous RGB-D sequences, we demonstrate that our method can promote performance over baselines. Our code, data, and model will be made publicly available.
Code: https://github.com/wuzirui/async-nerf.

\section{Related Works}
\label{sec:related}

\noindent\textbf{Large-scale Radiance Fields.} Neural Radiance Field (NeRF) \cite{mildenhall_nerf_2020} has shown impressive results in neural reconstruction and rendering. However, its capacity to model large-scale unbounded 3D scenes is limited. Several strategies \cite{turki_mega-nerf_2022, tancik2022block, xiangli2022bungeenerf, mi2023switchnerf, xu2023gridguided} have been proposed to address this limitation, with a common principle of dividing large scenes into blocks or decomposing the scene into multiple levels. Block-NeRF \cite{tancik2022block} clusters images by dividing the whole scene according to street blocks. Mega-NeRF \cite{turki_mega-nerf_2022} utilizes a clustering algorithm that partitions sampled 3D points into different NeRF submodules. BungeeNeRF \cite{xiangli2022bungeenerf} trains NeRFs using a growing model of residual blocks with predefined multiple scales of data. Switch-NeRF \cite{mi2023switchnerf} designs a gating network to jointly learn the scene decomposition and NeRFs without any priors of 3D scene shape or geometric distribution. However, these prior works fail to leverage the rich geometric information in depth images for effective regularization.

\noindent\textbf{Depth-regularized Radiance Fields.} Volumetric rendering requires extensive samples and sufficient views to effectively differentiate between empty space and opaque surfaces. Depth maps can serve as geometric cues, providing regularization constraints and sampling prior, which accelerates NeRF's convergence towards the correct geometry. DS-NeRF \cite{deng2022depth} enhances this process using depth supervision from 3D point clouds, estimated by structure-from-motion, and a specific loss for rendered ray termination distribution. Mono-SDF \cite{Yu2022MonoSDF} and Dense-Depth Prior \cite{roessle2022dense} further supplement this with a pretrained dense monocular depth estimator for less-observed and textureless areas. To adapt NeRF for outdoor scenarios, URF \cite{rematas2022urban} rasterizes a pre-built LiDAR point cloud map to generate dense depth images and alleviates floating elements by penalizing floaters in the free space. Moreover, S-NeRF \cite{ziyang2023snerf} completes depth on sparse LiDAR point clouds using a confidence map, effectively handling street-view scenes with limited perspectives. However, those methods are not readily applicable to UAV-captured images due to the lack of suitable synchronized sensors for long ranges.


\noindent\textbf{Broader UAV Vision and Synchronization.} Like autonomous driving, UAV vision is drawing increasing attention due to its unique characteristics. Broader UAV vision covers many topics like counting, trajectory forecasting, intention prediction, object tracking, physics understanding, next-best-view prediction, 3D reconstruction, calibration, and global camera localization \cite{zhu2023latitude, wu2022sc}.
Sensor synchronization is challenging for UAV vision (and other settings) and several works address the problem from an algorithmic perspective. One possibility is to adopt tailored hardware designs or software protocols \cite{ansari_wireless_2019} to synchronize all the devices. Another branch of sensor-agnostic methods utilizes temporal priors by using Sum-of-Gaussians \cite{elhayek_spatio-temporal_2012} or parametric interpolation functions \cite{yang_asynchronous_2021}.

\section{Preliminaries}
\label{sec:motivation}
\textbf{Problem \& Challenge.} Our goal is to learn a neural radiance field parameterized by \bm{$\theta$} for large-scale scene representation from UAV images as done in prior works \cite{turki_mega-nerf_2022,xiangli2022bungeenerf}. However, these prior works fail to leverage depth supervision, which is known \cite{deng2022depth, roessle2022dense} as useful for training floater-less NeRFs. To our knowledge, there are no easily accessible synchronized RGB-D sensor suites for \textbf{large-scale} outdoor scenes, and synchronizing them according to timestamp cannot fully address the misalignment issue. Instead of using expensive hardware, we take an algorithmic perspective. 

\textbf{Pose Representation.} We use a translation vector $\hat{\mathbf{x}} \in \mathbb{R}^3$ and a rotation matrix $\hat{\mathbf{q}} \in \rm{SO}(3)$ as our pose representation.

\noindent\textbf{Input \& output.} There are some prior works on large-scale scene modeling using aerial images \cite{xiangli2022bungeenerf, turki_mega-nerf_2022, guedonscone, guedon2023macarons}. In this study, we assume an input RGB-D stream captured by the drone: a set of RGB camera images $\{I^{(i)}\}_{i=1}^{N_I}$ and a set of depth maps $\{D^{(j)}\}_{j=1}^{N_D}$ shown in Fig.~\ref{fig:teaser}-a and we aim to recover the spatiotemporal transformations between them. Without loss of generality, we assume a set of camera poses $\{\mathcal{T}_I^{(i)}\}_{i=1}^{N_{I}}$  for color images are obtained by a Structure from Motion (SfM) algorithm. The neural scene representation parameterized by \bm{$\theta$} outputs an image $\hat{I}$ as well as a depth map $\hat{D}$ at a given perspective camera pose $\mathcal{T}_I$ and $\mathcal{T}_D$.




\begin{figure*}[ht]
\centering
\includegraphics[width=0.8\textwidth]{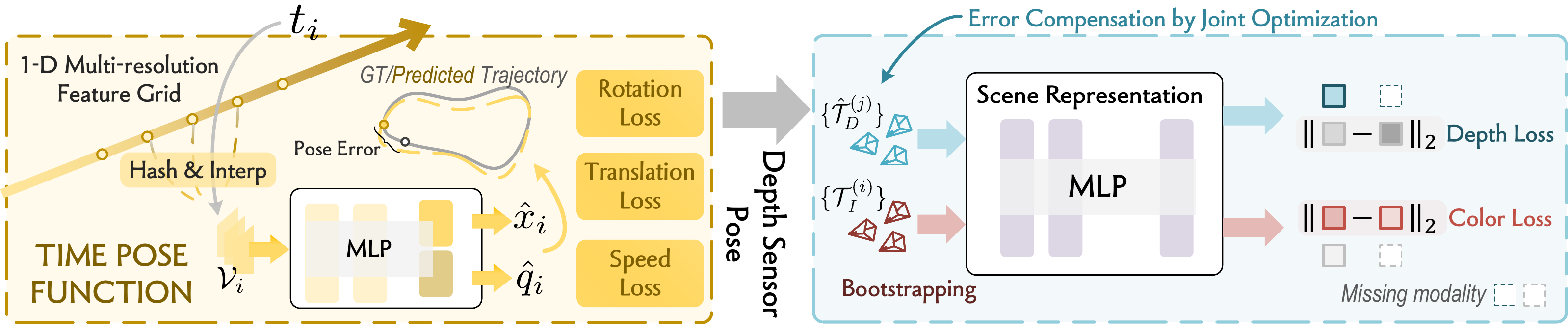}
\caption{\textbf{Method Pipeline.} The time-pose function is modeled using a 1-D multi-resolution hash grid with direct and speed losses. After bootstrapping the scene representation networks with pure RGB signals, the predicted depth sensor poses are used for jointly optimizing the NeRFs' parameters \bm{$\theta$}. At each timestamp ($t_i$ from RGB sequence or $t_j$ from depth sequence), only one modality of sensor signals is provided, thus only one loss term is activated (shown on the right).}
\label{fig:main-figure}
\end{figure*}

\section{Method}
\label{sec:method}

We introduce in Section \ref{subsec:method-timepose} the details of learning an implicit time-pose function. In Section \ref{sec:bootstrap}, we describe our neural scene representation networks and the bootstrapping strategy. In Section \ref{sec:joint-optim}, we adopt depth supervision and jointly train the time-pose function with RGB-D pairs. In Section \ref{sec: pipeline}, we summarize the overall pipeline of our method, for clarity.

\subsection{Time-Pose Function}
\label{subsec:method-timepose}
We represent the camera trajectory as an implicit time-pose function \bm{$\phi$} whose input is a timestamp, and whose output is a 6-DoF pose.

\textbf{Definition.} Specifically, the time-pose function can be represented as
    $\phi:t\to\hat{\mathcal T} = [\hat{\mathbf{x}},\hat{\mathbf{q}}]$,
where $t$ is the timestamp of capture, and $\hat{\mathcal T}$ is the estimated pose which is represented by a translation vector $\hat{\mathbf{x}} \in \mathbb{R}^3$ and a rotation matrix $\hat{\mathbf{q}} \in \rm{SO}(3)$.

\begin{figure}[ht]
\centering
\includegraphics[width=0.47\textwidth]{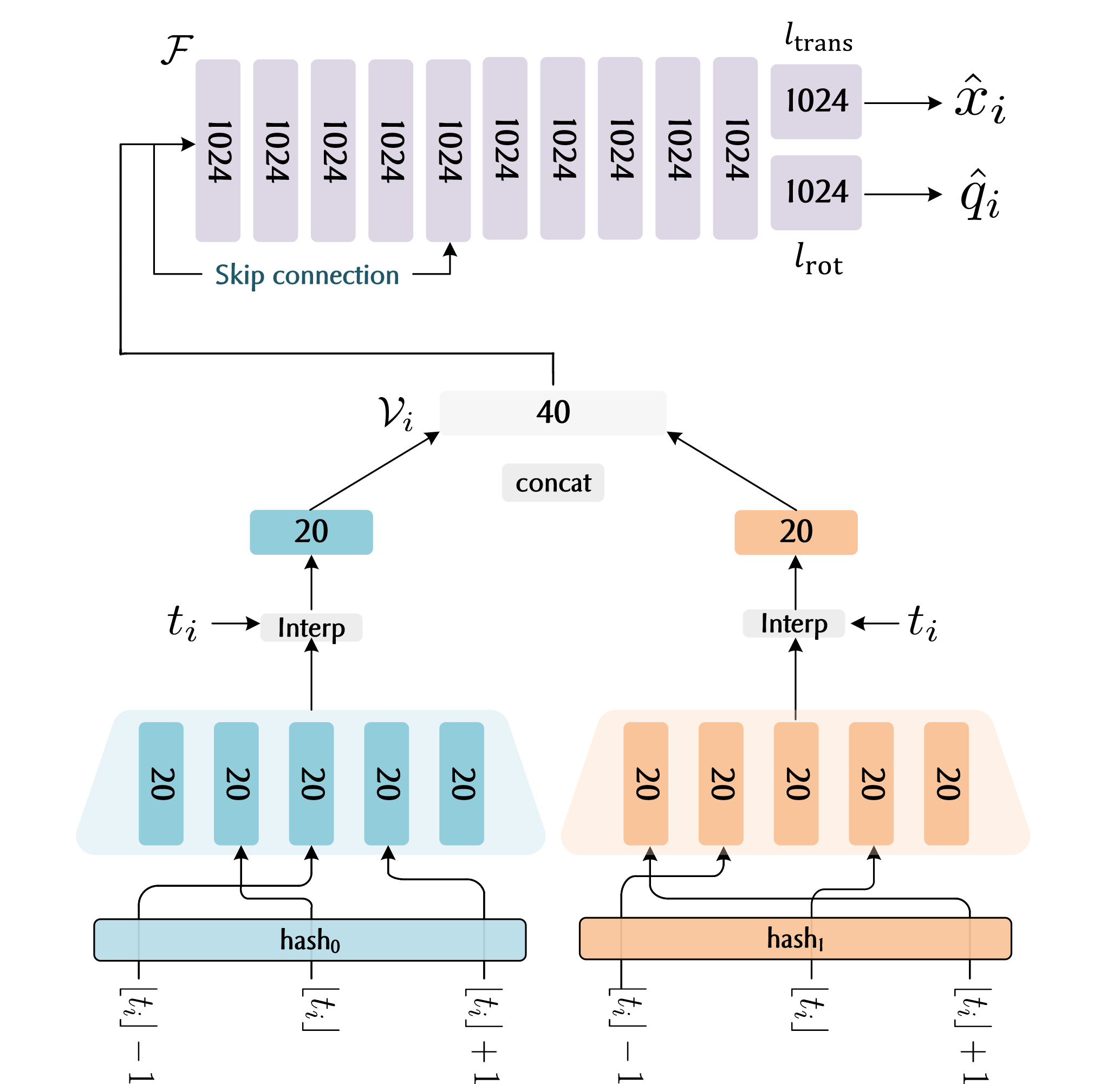}
\caption{Our implementation of the Time-Pose Function with a multi-resolution hash grid. Blue and orange networks are of different resolution.}
\label{fig:hash}
\end{figure}

We represent the time-pose function using a neural network. Fig.~\ref{fig:hash} depicts the network structure, comprising a compact 1-D multi-resolution hash grid ${\{\mathcal{G}^{(l)}}\}_{l=1}^L$ and an MLP decoder. 

\textbf{Hash encoding.} For the queried time-stamp $t_i$, the hash encoding of $\left\lfloor x_i \right\rfloor - 1,\left\lfloor x_i \right\rfloor$, and $\left\lfloor x_i \right\rfloor + 1$ are extracted in each layer $\mathcal G^{(l)}$ of the multi-resolution hash grid. 
We perform quadratic interpolation on the extracted hash feature:
\begin{equation}
    \begin{aligned}
        \mathcal V_i^l
        &= \frac 1 2(t_i - \left\lfloor x_i \right\rfloor) (t_i-\left\lfloor x_i \right\rfloor-1) \mathcal{G}^{(l)}(\left\lfloor x_i \right\rfloor-1)\\
        &-(t_i - \left\lfloor x_i \right\rfloor + 1) (t_i - \left\lfloor x_i \right\rfloor -1) \mathcal{G}^{(l)}(\left\lfloor x_i \right\rfloor)\\
        &+\frac 1 2 (t_i - \left\lfloor x_i \right\rfloor + 1) (t_i - \left\lfloor x_i \right\rfloor) \mathcal{G}^{(l)}(\left\lfloor x_i \right\rfloor + 1).
    \end{aligned}
\end{equation}
The interpolated feature are then concatenated together $\mathcal V_i = \text{concat}\{\mathcal V_i^l\}_{l=1}^L$.

\textbf{Decoding.} After obtaining the interpolated feature vector, an MLP with two separated decoder heads is used to predict the output translation $\hat x_i$ and rotation $\hat q_i$ vectors respectively. Specifically, we use a 10-layer MLP with 1024 dimensions in each layer, including a skip connection that concatenates the timestamp input to the \emph{5-th} layer which is shown in Fig.~\ref{fig:hash}.

The forward pass can be expressed in the following equations:
\begin{align}
    \mathcal V_i &= \text{concat}\{\text{interp}(\text{h}(t; \pi_l),\ \mathcal G^{(l)})\}_{l=1}^L\}, \\
    \mathcal{\hat T}_i &= [\hat x_i, \hat q_i] = l_\text{trans}(\mathcal{F}_\text{MLP}(\mathcal V_i)),\ l_\text{rot}(\mathcal{F}_\text{MLP}(\mathcal V_i)),
\end{align}
where $\text{interp}$ denotes interpolation, $\text{h}$ is the hash function parameterized by $\pi_l$, $\mathcal{F}_\text{MLP}, l_\text{trans}, l_\text{rot}$ are the MLP networks and the decoder heads, with $\Phi_\text{MLP}, \Phi_\text{trans}, \Phi_\text{rot}$ representing their parameters.

\textbf{Justification.} The reason why we choose this architecture is as follows: The time-pose function is a coordinate-based function that may contain coarse and fine-level feature components\footnote{This is shown by the ground truth trajectory in Fig.~\ref{fig:data-city}.}\cite{tancik_fourier_2020}, and this architecture allows us to sample the hash encodings from each grid layer with different resolutions and perform quadratic interpolation on the extracted encodings to obtain a feature vector $\mathcal V_i$ when querying a specific timestamp $t$ that is in the range of all timestamps.

\textbf{RGB-D Transformation.} Using the time-pose function, we can predict the camera pose for the depth images. Since both the depth maps and the RGB images are collected by the same drone, they cover the same spatial-temporal footprints except for the difference in the placement of the two sensors on the aircraft. For every depth frame, we first predict the RGB camera pose using the capture timestamps of the depth sensor with the time-pose function then transform the predicted RGB camera pose to the depth sensor pose with a pre-calibrated pose transformation $\mathcal{T}_{I\to D}$ between sensors.

\textbf{Training.} To train the Time-Pose Function, we propose the following objective function:
\begin{equation}
\mathcal{L}=\lambda_\text{trans}\mathcal{L}_\text{trans}+\lambda_\text{rot}\mathcal{L}_\text{rot}+\lambda_\text{speed}\mathcal{L}_\text{speed},
\end{equation}
where $\mathcal{L}_\text{trans},\mathcal{L}_\text{rot},\mathcal{L}_\text{speed}$ are translation, rotation and speed losses respectively as shown in the left panel of Fig.~\ref{fig:main-figure}. and $\lambda_\text{trans}, \lambda_\text{rot}, \lambda_\text{speed}$ are the weighting parameters. Note that $\lambda_\text{trans}$ and $\lambda_\text{rot}$ are automatically adjusted as explained in a later paragraph.

To optimize the translation and the rotation vectors, we minimize the mean square error (MSE) between the estimated and ground-truth camera poses:
\begin{equation}
    \mathcal{L}_\text{trans}=\frac 1 n \sum_{i=1}^n(x_i-\hat{x}_i)^2,\ 
    \mathcal{L}_\text{rot}=\frac 1 n \sum_{i=1}^n(q_i-\hat{q}_i)^2.
\end{equation}

\textbf{Uncertainty Balancing.} Since $x$ and $q$ are in different units, the scaling factor $\lambda_\text{trans}$ and $\lambda_\text{rot}$ play an important role in balancing the losses. To prevent translation and rotation from negatively influencing each other in training and to tap into possible mutual facilitation, we make the weighting factors learnable by using homoscedastic uncertainty \cite{kendall2017geometric} as $\mathcal{L}_\sigma=\mathcal{L}_\text{trans}\exp(-\hat{s}_\text{trans})+\hat{s}_\text{trans}+\mathcal{L}_\text{rot}\exp(-\hat{s}_\text{rot})+\hat{s}_\text{rot}$, where $\hat{s}$ are learnable parameters, thus the loss terms are balanced during training course\footnote{ Manual selection of weights requires laborious tuning, but comparable performance can be achieved.}.

\textbf{Speed.} Observing that the time-pose function is essentially a function of translational displacement and angular displacement with respect to time, we can use the average linear speed\footnote{Note that the average speed refers to the mean value calculated from the ground-truth camera pose of the current frame and the two adjacent frames, rather than the average value in the whole sequence.} to supervise the gradient of the network output, with regard to the input vectors. Since the linear speed variation is small and the angular speed variation is relatively larger in the scenes captured by the drone, only the average linear speed is used to supervise the neural network and the latter is not supervised in our method:
\begin{align}
    \mathcal{L}_\text{speed} &= \text{MSE}(v(t_i), \hat{v}(t_i)) \nonumber \\
    &= \frac{1}{n} \sum_{i=1}^n(v(t_i)-\frac{\partial \hat x}{\partial t}(t_i))^2 \\
    v(t_i) &= \left.\frac{\partial x}{\partial t}\right|_{t=t_i} \approx \frac{x_i-x_{i-1}}{t_i-t_{i-1}} \nonumber
\end{align}

\subsection{Bootstrapping Large-scale Neural Radiance Fields}
\label{sec:bootstrap}
\textbf{Partition.} In this part, we introduce our proposed scene representation shown in the right half of Fig.~\ref{fig:main-figure}.
Due to the limited capacity of MLPs,  we follow Mega-NeRF \cite{turki_mega-nerf_2022} and partition the scene map into a series of equal-sized blocks in terms of spatial scope, and each block learns its individual scene representation with an implicit field. In this stage, we optimize the scene representation with pure RGB data. Specifically, the radiance field is denoted as $\{f_\text{NeRF}^{(i) }\}_{i=1}^{N_x\times N_y}$, where $N_x, N_y$ denotes the spatial grid size.
Each implicit function represents a geographic region with $\textbf{r}_i^\text{centroid}$ as its centroid. The $k$-th scene model can be written as:
\begin{equation}
    f^\text{(k)}_\text{NeRF}(\gamma(\mathbf{r}_\text{pts}),\gamma(\mathbf{d}))\to(\hat c,\sigma),
\end{equation}
where k $= \mathop{\arg\min}\limits_{i}||\mathbf{r}_\text{pts}-\mathbf{r}_{i}^\text{centroid}||_2$ and $\gamma$ is the positional encoding function.

\textbf{Rendering.} For view synthesis, we adopt volume rendering techniques to synthesize color image $\hat{I}$ and depth map $\hat{D}$. Specifically, we sample a set of points for each emitted camera ray in a coarse-to-fine manner \cite{mildenhall_nerf_2020} and accumulate the radiance and the distance along the corresponding ray to calculate the rendered color $\hat{I}$ and depth $\hat{D}$.
To obtain the radiance of a spatial point $\mathbf{r}_\text{pts}$, we use the nearest scene model for prediction. A set of per-image appearance embedding \cite{martin-brualla_nerf_2021} is also optimized simultaneously in the training.

\begin{equation}
    \begin{split}
        {\hat I}(\mathbf o, \mathbf d) &= \int_\text{near}^\text{far}T(t)\sigma^{(k)}(\mathbf{r}(t))\cdot c^{(k)}(\mathbf{r}(t),\mathbf{d})\text dt,\ \\
        {\hat D}(\mathbf o, \mathbf d) &= \int_\text{near}^\text{far}T(t)\sigma^{(k)}(\mathbf{r}(t))\cdot t\text dt,
    \end{split}
\end{equation}
where $\textbf{o}$ and $\textbf{d}$ denote the position and orientation of the sampled ray, $\textbf{r}(t) = \textbf{o} + t\textbf{d}$ represents the sampled point coordinates in the world space, and $T(t)=\exp\left(-\int_\text{near}^t\sigma^{(k)}(\textbf{r}(s))\text{d}s\right)$ is the accumulated transmittance. We optimize the scene representation model with only the photometric error as $\mathcal{L}_\text{bootstrarp} = \text{MSE}(I, \hat{I})$. We empirically observe that this bootstrapping is critical to the challenging third stage which jointly learns $\theta$ and $\phi$ using asynchronous RGB-D data.

\subsection{Joint Optimization}
\label{sec:joint-optim}
While the time-pose function learns a good initialization from the RGB sequence, there are still errors to be compensated. In this section, we describe how we perform simultaneous mapping and pose optimization, which compensates for the initial error of the time-pose function. 

\textbf{Formulation.} We jointly optimize the inaccurate camera poses and the implicit maps:
when fitting parameters $\Theta_\text{NeRF}^{(k)}$ of the scene representation, the estimated depth camera poses $\hat{\mathcal{T}}_D^{(j)}\in \rm{SE}(3)$ (where $x\in \mathbb{R}^3$ and $q\in \rm{SO}(3)$) will be simultaneously optimized on the manifold:
\begin{equation}
    \theta,\{\hat{\mathcal{T}}_D\}=\underset{\theta, \mathcal{T} \in \rm{SE}(3)}{\operatorname{argmin}}  \mathcal{L}(\{I^{(i)}\}, \{D^{(j)}\}\mid \theta, \{\mathcal{\hat T}_D\}),
\end{equation}
where $\mathcal{L}$ is the objective function we demonstrate in the next paragraph, and the gradients on $\mathcal{T}$ can be further backpropagated to update $\phi$.

\textbf{Loss.} To train the implicit representation to obtain photo-realistic RGB rendering maps and accurate depth map estimation, we update the mapping losses as:
\begin{equation}
    \label{eq:barf_loss}
    \begin{split}
        \mathcal{L} = & \lambda_\text{color}\sum_{i}\text{MSE}(I^{(i)}, \hat{I}^{(i)}) \\
        & +\lambda_\text{depth}(\alpha)\sum_j\text{MSE}(D^{(j)}, \hat{D}^{(j)}),
    \end{split}
\end{equation}

where $\lambda_\text{color}$ and $\lambda_\text{depth}(\alpha)$ are weighting hyper-parameters for color and depth loss, in which the depth loss weight starts to grow from zero gradually with the training process.

To compensate for the error from the time-pose function extracted poses, we jointly optimize two implicit representation networks thanks to the end-to-end differentiable nature.

\begin{figure}[ht]
    \centering
    \includegraphics[width=0.45\textwidth]{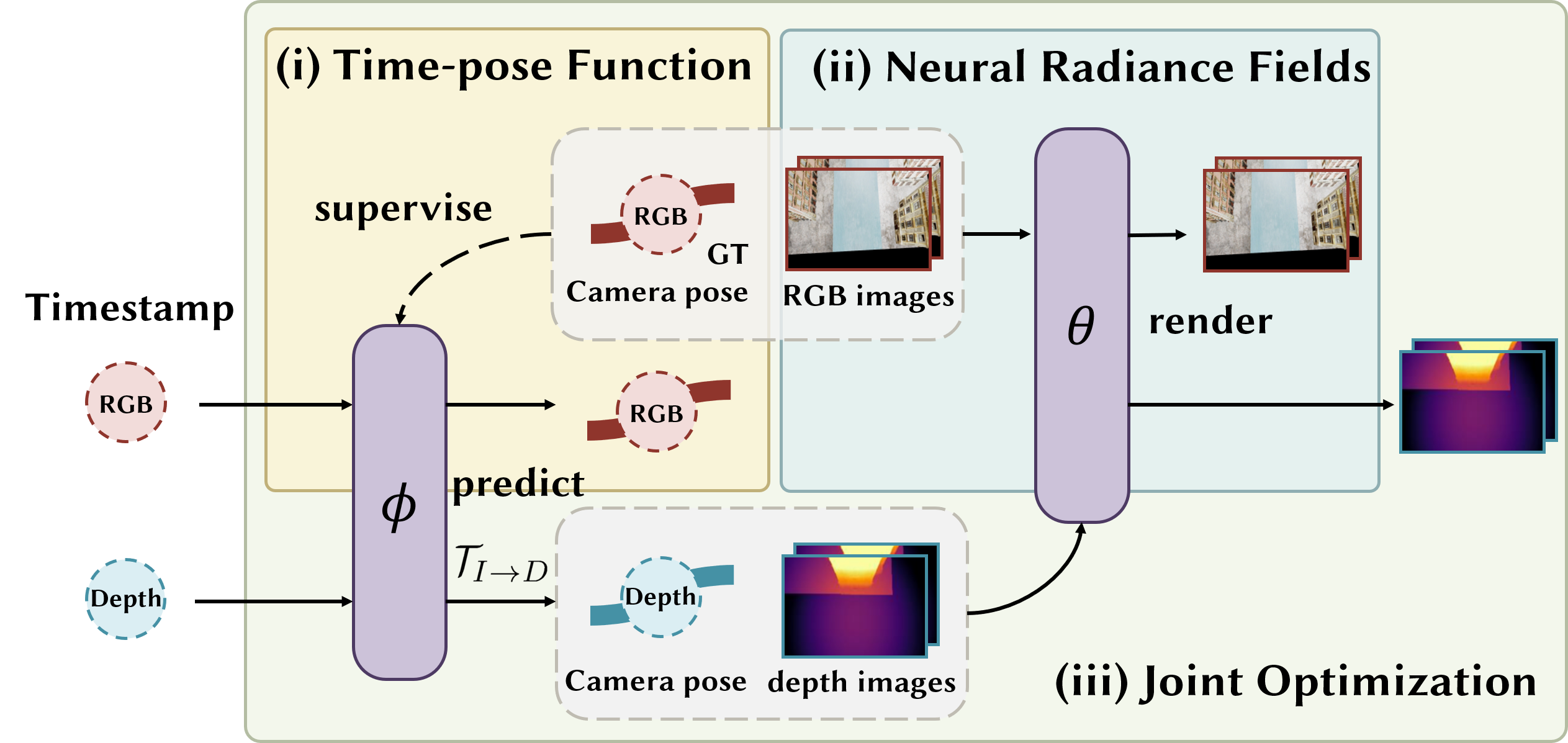}
    \caption{\textbf{Three-step Optimization.} (i) A time-pose function \bm{$\phi$} is trained to predict camera poses from timestamps; (ii) The neural radiance field parameterized by \bm{$\theta$} is bootstrapped with pure RGB losses; (iii) Both of the parameters \bm{$\theta$}, \bm{$\phi$} are jointly optimized with RGB-D supervision.}
    \label{fig:optimization_flow}
\end{figure}

\subsection{Pipeline Summary}
\label{sec: pipeline}
We propose a 3-step optimization as shown in Fig.~\ref{fig:optimization_flow}. First, since the time-pose relationship for the RGB captures is given, we can train a time-pose function on the RGB sequence. Then, to train the neural radiance field, we first bootstrap the network with pure RGB supervision. To further enable training with RGB-D supervision, we use the previously trained time-pose function and a pre-calibrated pose transformation $\mathcal{T}_{I\to D}$ to estimate the corresponding depth camera poses $\{\mathcal{T}_D^{(j)}\}$ of the depth timestamps $\{t_D^{(j)}\}$. Since both of the networks are differentiable, we jointly optimize the networks in an end-to-end manner in the third stage.

\begin{figure*}[ht]
\centering
\includegraphics[width=0.8\textwidth]{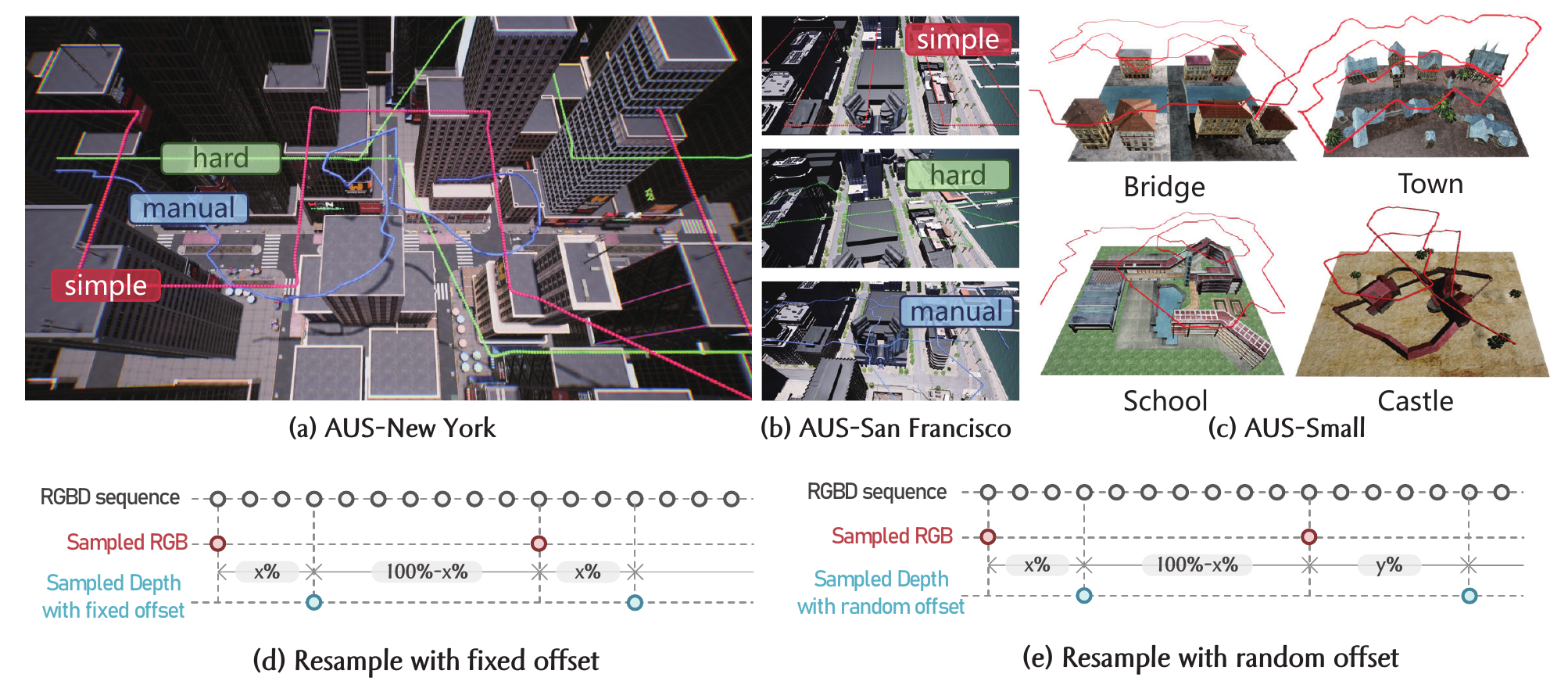}

\caption{We propose a photo-realistically rendered dataset named Asynchronous Urban Scene (AUS) for evaluation. (a/b) are large-scale city scenes designed according to New York and San Francisco while (c) is (relatively) small-scale scenes provided by UrbanScene3D. Drone trajectories of different difficulty levels are visualized in (a-c). On these trajectories, we first capture an RGB-D sequence with an enough high framerate. Then we exploit two resampling strategies: fixed offset (d) and random offset (e). $x$ equals $30$ in (d) for every RGB-D pair. $x$ equals $30$ while $y$ equals $50$ in (e).}
\label{fig:data-city}
\end{figure*}

\section{Experiments}
\label{sec:exp}

In Section \ref{sec: aus}, we introduce how we designed the AUS dataset. In Section \ref{subsec:results}, we conduct qualitative and quantitative evaluations of our proposed methods and compared them with baseline methods and demonstrate the effectiveness of our 3-step optimization pipeline. In Section \ref{sec: ablation}, we perform some ablation experiments.

\subsection{Asynchronous Urban Scene (AUS) Dataset}
\label{sec: aus}
\textbf{Dataset Collection.} Our Asynchronous Urban Scene (AUS) dataset as illustrated in Fig.~\ref{fig:data-city} is generated using Airsim \cite{shah2018airsim}, a simulator plug-in for Unreal Engine. With 3D city models loaded in Unreal Engine, the simulator can output photorealistic and high-resolution RGB images with synchronized depth images (resampled later) according to the a drone trajectory and a capture framerate. We choose Airsim as it strikes a good balance between rendering quality and dynamics modeling flexibility.

\textbf{3D City Scene Models.} To generate the AUS dataset, we exploit a total of six scene models, covering two large-scale ones shown in Fig.~\ref{fig:data-city}-a/b and four (relatively) small-scale ones shown in Fig.~\ref{fig:data-city}-c. The former uses the New York and San Francisco city scenes provided by Kirill Sibiriakov \cite{ArtStation}, in which AUS-NewYork covers a $ 250\times 150 m^{2}$ area with many detailed buildings and AUS-SanFrancisco consists of a $500\times 250 m^{2}$ area near the Golden Gate Bridge. The latter uses four model files provided in the UrbanScene3D dataset \cite{UrbanScene3D}. As such, at the scene level, AUS features a good coverage of both large-scale modern cities and smaller cultural heritage sites.

\textbf{Trajectory Generation.} Trajectory complexity matters for our problem. In many real-world applications, photographers may manually control drones to capture a city. To build a meaningful and comprehensive benchmark, we use three types of trajectories: a trivial Zig-Zag trajectory named simple, a more complex randomly generated trajectory named hard, and a very complex manually controlled trajectory named manual in Fig.~\ref{fig:data-city}. In AUS-Small, we only provide manually controlled trajectories, since the scene sizes are relatively small and using the former two trajectory strategies leads to an unrealistically large overlap between frames.

\textbf{Mismatch Resampling.} We first sample synchronized RGB-D sequences in the simulator at a high frequency (50fps) then re-sample RGB and depth images with various offsets to create asynchronous RGB-D sequences. As shown in Fig.~\ref{fig:data-city}-d/e, we exploit two settings for the AUS dataset. In Fig.~\ref{fig:data-city}-d, every RGB-D pair is resampled according to a fixed offset denoted by percent $\rm x$. For example, we sample the RGB image at 5fps or say every 10 frames and $\rm x=30$ means every depth image is 3 frames later than the RGB counterpart. In Fig.~\ref{fig:data-city}-e, the offset between an RGB-D pair is randomly selected, simulating a challenging real-world asynchronous sequence. Offset ablation will be shown later.

\begin{figure}[t]
\centering
\includegraphics[width=0.49\textwidth]{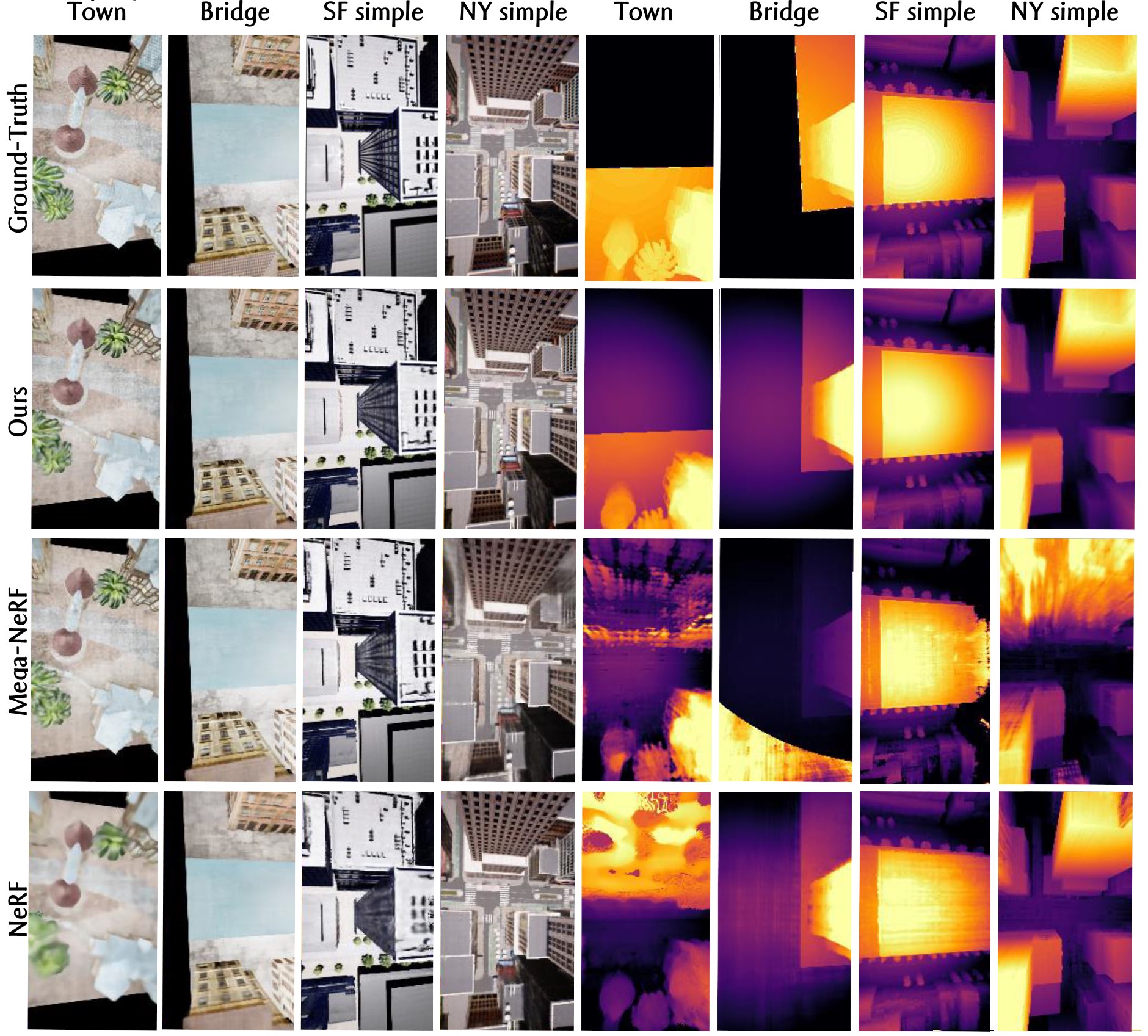} 
\caption{\textbf{Qualitative Results.} Due to the large size of the dataset, we have selected a few representative samples to showcase our results. It can be seen that our method can render photo-realistic novel views (left) and the best depth estimation results (right). Please zoom in to see improvements.}
\label{fig:qualitative}
\end{figure}

\subsection{Comparison to State-of-the-art}
\label{subsec:results}

We evaluate our proposed method against NeRF-W \cite{martin-brualla_nerf_2021}, city-scale Mega-NeRF \cite{turki_mega-nerf_2022}, linear interpolation and SIREN. The latter two are variants of our 1D hash-function. We present the quantitative results in Table.~\ref{tab:rgbd-quantitative} and qualitative results in Fig. \ref{fig:qualitative}. NeRF-W is the baseline from which we borrow the aforementioned idea of per-image appearance embedding and Mega-NeRF is a large-scale scene modeling framework which our network is built upon. Linear interpolation and SIREN~\cite{sitzmann2019siren} are two alternative approaches that can be used to estimate camera pose from timestamps.

\textbf{RGB-D View Synthesis}
We present the RGB-D view synthesis results qualitatively in Fig.~\ref{fig:qualitative} in which our method synthesize photo-realistic images and accurate depth maps, while baseline methods fail at predicting reliable depth maps (e.g., in NY hard, the depth values around glasses are obviously inaccurate). For quantitative comparison (Table.~\ref{tab:rgbd-quantitative}), the standard metrics for novel view synthesis and depth estimation are used for evaluation. For RGB view synthesis, image quality metrics including PSNR, SSIM\cite{1284395}, and the VGG implementation of LPIPS \cite{zhang_unreasonable_2018} are used. The Peak Signal-to-Noise Ratio (PSNR) is in inverse ratio to the mean square error. Our method achieves a higher PSNR than other methods, with an improvement of 1.11 dB over Mega-NeRF. In addition to PSNR, the Structural Similarity Index (SSIM) is designed to model image distortion as a combination of three factors: structure, luminance, and contrast. As it can be seen, the SSIM value obtained by our method is the highest among all images, reaching 0.8206. The Learned Perceptual Image Patch Similarity (LPIPS) metric focuses more on perceptual factors that are relevant to human vision, allowing for better capturing of perceptual differences between images. The LPIPS score typically ranges from 0 to 1, where a smaller value indicates higher perceptual quality of the image. Our method has achieved the minimum LPIPS score. Therefore, the results indicate that our method achieves a higher image quality.




For depth estimation, Root Mean Square Error (RMSE), Root Mean Square Logarithmic Error (RMSE log), $\delta_{\text{thr}} = \frac{1}{MN}\sum_{i,j} (\max(\frac{f_{ij}} {g_{ij}}, \frac{g_{ij}}{f_{ij}}) < 1.25^{\text{thr}})$ are used, where $f$ and $g$ are $M*N$ depth images. Our method achieves a significant reduction in RMSE by 35.26 and a decrease in RMSE log by 0.3332. Additionally, it demonstrates an improvement in the $\delta_{1,2,3}$ metric, with an average enhancement around 12\%.

\begin{table*}[t]
\centering
\caption{Evaluation results on the AUS dataset. For the NY and SF, we report the mean performances on all sequences (Simple / Hard / Manual). For the Bridge / Town / School / Castle scenes, we integrate their results and presented them as the "Small". }
\begin{tabular}{cccccccccc}
\hline
\textbf{Scene} & \textbf{Method} & PSNR $\uparrow$ & SSIM $\uparrow$ & LPIPS $\downarrow$ & RMSE $\downarrow$ & RMSE log $\downarrow$ & $\delta_1 (\%)\ \uparrow$ & $\delta_2 (\%)\ \uparrow$ & $\delta_3 (\%)\ \uparrow$ \\ \hline
 & NeRF-W & 23.32 & 0.8105 & 0.2249 & 17.40 & 0.2630 & 80.17 & 90.11 & 94.72 \\
 & Mega-NeRF & 23.53 & \textbf{0.8375} & 0.1920 & 23.99 & 0.2943 & 80.11 & 88.77 & 92.87 \\
 & Linear Interp. & 21.12 & 0.8034 & 0.3065 & 27.14 & 0.4643 & 64.24 & 83.12 & 88.72 \\
 & SIREN & 20.78 & 0.8012 & 0.3543 & 32.45 & 0.5012 & 58.97 & 81.50 & 87.80 \\
\multirow{-5}{*}{\begin{tabular}[c]{@{}c@{}}\textbf{NY}\\ Mean\end{tabular}} & Ours & \textbf{24.33} & 0.8346 & \textbf{0.1833} & \textbf{6.15} & \textbf{0.0816} & \textbf{94.85} & \textbf{98.23} & \textbf{99.22} \\ \hline
 & NeRF-W & 19.21 & 0.6610 & 0.3632 & 24.93 & 0.1877 & 81.81 & 91.54 & 96.93 \\
 & Mega-NeRF & 20.53 & 0.7334 & \textbf{0.2619} & 23.56 & 0.1713 & 88.58 & 94.74 & 96.83 \\
 & Linear Interp. & 18.01 & 0.6514 & 0.4576 & 30.84 & 0.2501 & 72.38 & 87.38 & 95.32 \\
 & SIREN & 18.22 & 0.6423 & 0.4462 & 28.49 & 0.2379 & 74.01 & 88.08 & 95.61 \\
\multirow{-5}{*}{\begin{tabular}[c]{@{}c@{}}\textbf{SF}\\ Mean\end{tabular}} & Ours & \textbf{22.14} & \textbf{0.7930} & 0.2620 & \textbf{7.64} & \textbf{0.0789} & \textbf{96.34} & \textbf{98.80} & \textbf{99.70} \\ \hline
 & NeRF-W & 22.61 & 0.6855 & 0.3384 & 99.90 & 1.1233 & 59.03 & 67.27 & 68.45 \\
 & Mega-NeRF & 26.58 & 0.8193 & 0.2250 & 92.00 & 1.0326 & 70.13 & 75.08 & 76.29 \\
 & Linear Interp. & 21.34 & 0.6586 & 0.3575 & 128.44 & 1.1585 & 44.12 & 55.93 & 60.27 \\
 & SIREN & 22.01 & 0.6345 & 0.3571 & 132.76 & 1.1298 & 47.94 & 58.34 & 64.83 \\
\multirow{-5}{*}{\textbf{Small}} & Ours & \textbf{27.53} & \textbf{0.8343} & \textbf{0.2218} & \textbf{20.00} & \textbf{0.3382} & \textbf{92.79} & \textbf{96.56} & \textbf{98.18} \\ \hline

\rowcolor[HTML]{D6D6D6} 
\cellcolor[HTML]{D6D6D6} & NeRF-W & 21.71 & 0.7190 & 0.3088 & 47.41 & 0.5247 & 73.67 & 82.97 & 86.70 \\
\rowcolor[HTML]{D6D6D6} 
\cellcolor[HTML]{D6D6D6} & Mega-NeRF & 23.55 & 0.7967 & 0.2263 & 46.52 & 0.4994 & 79.61 & 86.20 & 88.66 \\
\rowcolor[HTML]{D6D6D6} 
\cellcolor[HTML]{D6D6D6} & Linear Interp. & 20.16 & 0.7012 & 0.3739 & 62.14 & 0.6243 & 60.25 & 75.48 & 81.44 \\
\rowcolor[HTML]{D6D6D6} 
\cellcolor[HTML]{D6D6D6} & SIREN & 20.34 & 0.6926 & 0.3859 & 64.56 & 0.6230 & 60.31 & 75.97 & 82.75 \\
\rowcolor[HTML]{D6D6D6} 
\cellcolor[HTML]{D6D6D6} & \cellcolor[HTML]{D6D6D6} & \textbf{24.66} & \textbf{0.8206} & \textbf{0.2224} & \textbf{11.26} & \textbf{0.1662} & \textbf{94.66} & \textbf{97.86} & \textbf{99.03} \\
\rowcolor[HTML]{D6D6D6} 
\multirow{-4}{*}{\cellcolor[HTML]{D6D6D6}\textbf{Mean}} & \multirow{-2}{*}{\cellcolor[HTML]{D6D6D6}Ours} & {\color[HTML]{FE0000} (+1.11)} & {\color[HTML]{FE0000} (+0.0239)} & {\color[HTML]{FE0000} (-0.0039)} & {\color[HTML]{FE0000} (-35.26)} & {\color[HTML]{FE0000} (-0.3332)} & {\color[HTML]{FE0000} (+15.05)} & {\color[HTML]{FE0000} (+11.66)} & {\color[HTML]{FE0000} (+10.37)} \\ \hline
\end{tabular}%
\label{tab:rgbd-quantitative}
\end{table*}

\begin{table}[th]
\centering
\caption{Performance of the time-pose function. For large cities, results for different trajectories are separated by slashes. For small scenes, results for different scenes are separated by slashes.}

\resizebox{0.48\textwidth}{!}{
\begin{tabular}{lll}
\hline
\multicolumn{1}{c}{\multirow{2}{*}{\textbf{Scene}}} &
\multicolumn{2}{c}{\textbf{Time-Pose Function}}\\
\multicolumn{1}{c}{} &
  Rotation ($^\circ$) &
  Translation ($m$) \\ \hline
\textbf{NY Full} & 0.66 / 0.59 / 3.70 & 1.84 / 1.12 / 0.46\\
\textbf{SF Full} & 0.17 / 0.67 / 0.65 & 1.34 / 1.45 / 0.94\\
\textbf{Small}   & 1.51 / 0.68 / 0.70 / 1.05 & 0.95 / 1.35 / 0.89 / 0.38 \\ \hline
\textbf{Mean} &
  \textbf{1.04} &
  \textbf{1.07} \\ \hline
  
  \multicolumn{1}{c}{\multirow{2}{*}{\textbf{Scene}}} &
  \multicolumn{2}{c}{\textbf{Joint Optimization}} \\
\multicolumn{1}{c}{} &
  Rotation ($^\circ$) &
  Translation ($m$) \\ \hline
\textbf{NY Full}& 0.13 / 0.09 / 1.47 & 0.34 / 0.56 / 0.20\\
\textbf{SF Full} & 0.05 / 0.41 / 0.02 & 0.32 / 1.09 / 0.66 \\
\textbf{Small} & 0.49 / 0.36 / 0.68 / 0.38 & 0.57 / 0.85 / 0.56 / 0.12 \\ \hline
\textbf{Mean} &
  \textbf{0.41 \textcolor{red}{(-0.63)}} &
  \textbf{0.53 \textcolor{red}{(-0.54)}} \\ \hline
\end{tabular}%
}
\label{tab:pose-quant}
\end{table}

\textbf{Depth Pose Estimation}
We evaluate the performance of our time-pose function (Table. ~\ref{tab:pose-quant}) to localize depth sensor poses. As shown quantitatively, our method can achieve an average pose error of $1.04m$ and $1.07^\circ$ in the first stage. After joint optimization in the third stage, our method cuts half the errors to $0.53m$ and $0.41^\circ$. The results show that the time-pose function learns an accurate implicit trajectory from the RGB sequence that can estimate accurate poses for depth frames. By further tuning the time-pose function jointly with the scene representation network, the accuracy of the predicted depth sensor poses can be improved.

\textbf{Real-world Evaluation.}
In the real-world experiments, we use the DJI M300 UAV (equipped with a high-definition RGB camera and LiDAR to collect real data, where the RGB camera collects images at the frame rate of 30fps and the LiDAR collects depth information at 240Hz. The poses of the RGB images are provided by COLMAP \cite{schonberger_structure--motion_2016}. The fixed transformations between sensors are provided by the producer. A qualitative comparison is provided in Fig.~\ref{fig:teaser}. In terms of depth rendering, the figure shows that our method outperforms Mega-NeRF in generating better depth maps.



\subsection{Ablation Studies}
\label{sec: ablation}

\begin{table}[t]
\centering
\caption{Ablation Study on the joint optimization stage.}
\resizebox{0.48\textwidth}{!}{%
\begin{tabular}{lllllll}
\hline
\multicolumn{1}{c}{\multirow{2}{*}{\textbf{Scene}}} & \multicolumn{2}{c}{\textbf{Ours}} & \multicolumn{2}{c}{\textbf{w/o depth input}} & \multicolumn{2}{c}{\textbf{w/o joint optimization}} \\ \cline{2-7} 
\multicolumn{1}{c}{} & PSNR $\uparrow$    & RMSE $\downarrow$    & PSNR $\uparrow$    & RMSE $\downarrow$    & PSNR  $\uparrow$   & RMSE $\downarrow$   \\ \hline
\textbf{NY} Mean                   & \textbf{24.24}   & \textbf{5.93}    & 24.03  & 42.15   & 19.70  & 15.94   \\
\textbf{SF} Mean                  & \textbf{22.70} & \textbf{7.26}    & 20.00 & 32.17 & 19.07 & 11.39 \\
\textbf{Bridge}               & \textbf{29.06} & \textbf{26.55}   & 27.98 & 120.41  & 22.35 & 96.16   \\
\textbf{Town}                 & \textbf{25.32}  & \textbf{15.61}   & 24.69 & 129.50   & 20.14  & 81.99   \\
\textbf{School}               & \textbf{26.51}   & \textbf{21.19}  & 25.57 & 63.10 & 21.91 & 42.74   \\
\textbf{Castle}               & \textbf{28.22} & \textbf{16.66} & 28.06 & 54.99 & 23.23   & 38.90 \\
\rowcolor[HTML]{D6D6D6}
\cellcolor[HTML]{D6D6D6}
\textbf{Mean}                 & \textbf{26.01} & \textbf{15.53} & 25.01 & 73.72 & 21.07   & 47.85 \\ \hline
\end{tabular}%
}
\label{tab:ablation-mega-nerf-depth}
\end{table}

\textbf{Joint Optimization for Pose Error Compensation.}
To demonstrate the importance of rectifying erroneous poses of depth images in asynchronous RGB-D sequences using the time-pose function, we train a Mega-NeRF~\cite{turki_mega-nerf_2022} with depth supervision but disabled the joint optimization stage. From the evaluation results (Table.~\ref{tab:ablation-mega-nerf-depth}), we observe its substantial impact on the rendering quality (PSNR for RGB and RMSE for depth), which shows that jointly optimizing the time-pose function and the scene representation significantly helps reduce geometric error. We also provide the setting \emph{without depth input}, which shows better than PSNR than \emph{w/o joint optimization} but much worse RMSE than it.

\textbf{Time-pose function.} We perform an ablation study with time-pose function (Table.~\ref{tab:rgbinit}). As a comparison, we use the pose from RGB images as the initial value of the depth camera pose and perform joint optimization in sequence NewYork-Hard. The results show that our method achieves better results by using the initial values generated by the time-pose function rather than directly using the transformed poses from the RGB sequence. The RGB poses are effective initial values when the drone motion is small, such as small rotations or slow movements. However, their performance deteriorates when the motion in the sequence becomes larger.

\begin{table}[h]
\centering
\caption{Ablation study on the Time-Pose Function.}
\resizebox{0.47\textwidth}{!}{%
\begin{tabular}{lllll}
\hline
 & \textbf{Rotation($^\circ$)} & \textbf{Translation(m)} & \textbf{PSNR} & \textbf{RMSE} \\ \hline
RGB Init. & 0.72 & 1.38 & 20.41 & 12.04 \\
Ours full & \textbf{0.09} & \textbf{0.56} & \textbf{24.33} & \textbf{6.15} \\ \hline
\end{tabular}%
}
\label{tab:rgbinit}
\end{table}

\section{Conclusion}
In this paper, we present a method to learn depth-supervised neural radiance fields from asynchronous RGB-D sequences. We leverage an important prior that the sensors cover the same spatial-temporal footprints and propose to utilize this prior with an implicit time-pose function. With a 3-staged optimization pipeline, our method calibrates the RGB-D poses and trains a large-scale implicit scene representation. Our experiments on a newly proposed large-scale dataset show that our method can effectively register depth camera poses and learns the 3D scene representation for photo-realistic novel view synthesis and accurate depth estimations. The future work includes leveraging the imagery features in the time-pose function and add more scenes into the AUS Dataset.









\bibliographystyle{IEEEtran}
\bibliography{IEEEabrv,iros}

\begin{thebibliography}{10}
\providecommand{\url}[1]{#1}
\csname url@rmstyle\endcsname
\providecommand{\newblock}{\relax}
\providecommand{\bibinfo}[2]{#2}
\providecommand\BIBentrySTDinterwordspacing{\spaceskip=0pt\relax}
\providecommand\BIBentryALTinterwordstretchfactor{4}
\providecommand\BIBentryALTinterwordspacing{\spaceskip=\fontdimen2\font plus
\BIBentryALTinterwordstretchfactor\fontdimen3\font minus \fontdimen4\font\relax}
\providecommand\BIBforeignlanguage[2]{{%
\expandafter\ifx\csname l@#1\endcsname\relax
\typeout{** WARNING: IEEEtran.bst: No hyphenation pattern has been}%
\typeout{** loaded for the language `#1'. Using the pattern for}%
\typeout{** the default language instead.}%
\else
\language=\csname l@#1\endcsname
\fi
#2}}

\bibitem{deng2022deep}
H.~Deng, M.~Bui, N.~Navab, L.~Guibas, S.~Ilic, and T.~Birdal, ``Deep bingham networks: Dealing with uncertainty and ambiguity in pose estimation,'' \emph{International Journal of Computer Vision}, pp. 1--28, 2022.

\bibitem{roessle2022dense}
B.~Roessle, J.~T. Barron, B.~Mildenhall, P.~P. Srinivasan, and M.~Nie{\ss}ner, ``Dense depth priors for neural radiance fields from sparse input views,'' in \emph{Proceedings of the IEEE/CVF Conference on Computer Vision and Pattern Recognition}, 2022, pp. 12\,892--12\,901.

\bibitem{xiangli2022bungeenerf}
Y.~Xiangli, L.~Xu, X.~Pan, N.~Zhao, A.~Rao, C.~Theobalt, B.~Dai, and D.~Lin, ``Bungeenerf: Progressive neural radiance field for extreme multi-scale scene rendering,'' in \emph{The European Conference on Computer Vision (ECCV)}, 2022.

\bibitem{rematas2022urban}
K.~Rematas, A.~Liu, P.~P. Srinivasan, J.~T. Barron, A.~Tagliasacchi, T.~Funkhouser, and V.~Ferrari, ``Urban radiance fields,'' in \emph{Proceedings of the IEEE/CVF Conference on Computer Vision and Pattern Recognition}, 2022, pp. 12\,932--12\,942.

\bibitem{wu2023mars}
Z.~Wu, T.~Liu, L.~Luo, Z.~Zhong, J.~Chen, H.~Xiao, C.~Hou, H.~Lou, Y.~Chen, R.~Yang, Y.~Huang, X.~Ye, Z.~Yan, Y.~Shi, Y.~Liao, and H.~Zhao, ``Mars: An instance-aware, modular and realistic simulator for autonomous driving,'' \emph{CICAI}, 2023.

\bibitem{mildenhall_nerf_2020}
\BIBentryALTinterwordspacing
B.~Mildenhall, P.~P. Srinivasan, M.~Tancik, J.~T. Barron, R.~Ramamoorthi, and R.~Ng, ``\BIBforeignlanguage{en}{{NeRF}: {Representing} {Scenes} as {Neural} {Radiance} {Fields} for {View} {Synthesis}},'' in \emph{\BIBforeignlanguage{en}{Computer {Vision} – {ECCV} 2020}}, ser. Lecture {Notes} in {Computer} {Science}, A.~Vedaldi, H.~Bischof, T.~Brox, and J.-M. Frahm, Eds.\hskip 1em plus 0.5em minus 0.4em\relax Cham: Springer International Publishing, 2020, pp. 405--421. [Online]. Available: \url{https://arxiv.org/abs/2003.08934}
\BIBentrySTDinterwordspacing

\bibitem{wang2021nerf}
Z.~Wang, S.~Wu, W.~Xie, M.~Chen, and V.~A. Prisacariu, ``Nerf--: Neural radiance fields without known camera parameters,'' \emph{arXiv preprint arXiv:2102.07064}, 2021.

\bibitem{lin2021barf}
C.-H. Lin, W.-C. Ma, A.~Torralba, and S.~Lucey, ``Barf: Bundle-adjusting neural radiance fields,'' in \emph{Proceedings of the IEEE/CVF International Conference on Computer Vision}, 2021, pp. 5741--5751.

\bibitem{jeong2021self}
Y.~Jeong, S.~Ahn, C.~Choy, A.~Anandkumar, M.~Cho, and J.~Park, ``Self-calibrating neural radiance fields,'' in \emph{Proceedings of the IEEE/CVF International Conference on Computer Vision}, 2021, pp. 5846--5854.

\bibitem{turki_mega-nerf_2022}
H.~Turki, D.~Ramanan, and M.~Satyanarayanan, ``Mega-{NeRF}: {Scalable} {Construction} of {Large}-{Scale} {NeRFs} for {Virtual} {Fly}- {Throughs},'' in \emph{2022 {IEEE}/{CVF} {Conference} on {Computer} {Vision} and {Pattern} {Recognition} ({CVPR})}, June 2022, pp. 12\,912--12\,921, iSSN: 2575-7075.

\bibitem{tancik2022block}
M.~Tancik, V.~Casser, X.~Yan, S.~Pradhan, B.~Mildenhall, P.~P. Srinivasan, J.~T. Barron, and H.~Kretzschmar, ``Block-nerf: Scalable large scene neural view synthesis,'' in \emph{Proceedings of the IEEE/CVF Conference on Computer Vision and Pattern Recognition}, 2022, pp. 8248--8258.

\bibitem{mi2023switchnerf}
\BIBentryALTinterwordspacing
Z.~Mi and D.~Xu, ``Switch-nerf: Learning scene decomposition with mixture of experts for large-scale neural radiance fields,'' in \emph{International Conference on Learning Representations (ICLR)}, 2023. [Online]. Available: \url{https://openreview.net/forum?id=PQ2zoIZqvm}
\BIBentrySTDinterwordspacing

\bibitem{xu2023gridguided}
L.~Xu, Y.~Xiangli, S.~Peng, X.~Pan, N.~Zhao, C.~Theobalt, B.~Dai, and D.~Lin, ``Grid-guided neural radiance fields for large urban scenes,'' 2023.

\bibitem{deng2022depth}
K.~Deng, A.~Liu, J.-Y. Zhu, and D.~Ramanan, ``Depth-supervised nerf: Fewer views and faster training for free,'' in \emph{Proceedings of the IEEE/CVF Conference on Computer Vision and Pattern Recognition}, 2022, pp. 12\,882--12\,891.

\bibitem{Yu2022MonoSDF}
Z.~Yu, S.~Peng, M.~Niemeyer, T.~Sattler, and A.~Geiger, ``Monosdf: Exploring monocular geometric cues for neural implicit surface reconstruction,'' \emph{Advances in Neural Information Processing Systems (NeurIPS)}, 2022.

\bibitem{ziyang2023snerf}
Z.~Xie, J.~Zhang, W.~Li, F.~Zhang, and L.~Zhang, ``S-ne{RF}: Neural radiance fields for street views,'' in \emph{The Eleventh International Conference on Learning Representations}, 2023.

\bibitem{zhu2023latitude}
Z.~Zhu, Y.~Chen, Z.~Wu, C.~Hou, Y.~Shi, C.~Li, P.~Li, H.~Zhao, and G.~Zhou, ``Latitude: Robotic global localization with truncated dynamic low-pass filter in city-scale nerf,'' in \emph{2023 IEEE International Conference on Robotics and Automation (ICRA)}.\hskip 1em plus 0.5em minus 0.4em\relax IEEE, 2023, pp. 8326--8332.

\bibitem{wu2022sc}
X.~Wu, H.~Zhao, S.~Li, Y.~Cao, and H.~Zha, ``Sc-wls: Towards interpretable feed-forward camera re-localization,'' in \emph{European Conference on Computer Vision}.\hskip 1em plus 0.5em minus 0.4em\relax Springer, 2022, pp. 585--601.

\bibitem{ansari_wireless_2019}
S.~Ansari, N.~Wadhwa, R.~Garg, and J.~Chen, ``Wireless {Software} {Synchronization} of {Multiple} {Distributed} {Cameras},'' in \emph{2019 {IEEE} {International} {Conference} on {Computational} {Photography} ({ICCP})}, May 2019, pp. 1--9.

\bibitem{elhayek_spatio-temporal_2012}
A.~Elhayek, C.~Stoll, N.~Hasler, K.~I. Kim, H.-P. Seidel, and C.~Theobalt, ``Spatio-temporal motion tracking with unsynchronized cameras,'' in \emph{2012 {IEEE} {Conference} on {Computer} {Vision} and {Pattern} {Recognition}}, June 2012, pp. 1870--1877, iSSN: 1063-6919 CCF: A.

\bibitem{yang_asynchronous_2021}
\BIBentryALTinterwordspacing
A.~J. Yang, C.~Cui, I.~A. Barsan, R.~Urtasun, and S.~Wang, ``Asynchronous {Multi}-{View} {SLAM},'' in \emph{2021 {IEEE} {International} {Conference} on {Robotics} and {Automation} ({ICRA})}, May 2021, pp. 5669--5676, iSSN: 2577-087X. [Online]. Available: \url{https://ieeexplore.ieee.org/document/9561481/}
\BIBentrySTDinterwordspacing

\bibitem{guedonscone}
A.~Guedon, P.~Monasse, and V.~Lepetit, ``Scone: Surface coverage optimization in unknown environments by volumetric integration,'' in \emph{Advances in Neural Information Processing Systems}, 2022.

\bibitem{guedon2023macarons}
A.~Gu{\'e}don, T.~Monnier, P.~Monasse, and V.~Lepetit, ``Macarons: Mapping and coverage anticipation with rgb online self-supervision,'' \emph{arXiv preprint arXiv:2303.03315}, 2023.

\bibitem{tancik_fourier_2020}
\BIBentryALTinterwordspacing
M.~Tancik, P.~P. Srinivasan, B.~Mildenhall, S.~Fridovich-Keil, N.~Raghavan, U.~Singhal, R.~Ramamoorthi, J.~T. Barron, and R.~Ng, ``Fourier {Features} {Let} {Networks} {Learn} {High} {Frequency} {Functions} in {Low} {Dimensional} {Domains},'' June 2020, arXiv:2006.10739 [cs]. [Online]. Available: \url{http://arxiv.org/abs/2006.10739}
\BIBentrySTDinterwordspacing

\bibitem{kendall2017geometric}
A.~Kendall and R.~Cipolla, ``Geometric loss functions for camera pose regression with deep learning,'' in \emph{Proceedings of the IEEE conference on computer vision and pattern recognition}, 2017, pp. 5974--5983.

\bibitem{martin-brualla_nerf_2021}
R.~Martin-Brualla, N.~Radwan, M.~S.~M. Sajjadi, J.~T. Barron, A.~Dosovitskiy, and D.~Duckworth, ``{NeRF} in the {Wild}: {Neural} {Radiance} {Fields} for {Unconstrained} {Photo} {Collections},'' in \emph{2021 {IEEE}/{CVF} {Conference} on {Computer} {Vision} and {Pattern} {Recognition} ({CVPR})}, June 2021, pp. 7206--7215, iSSN: 2575-7075.

\bibitem{shah2018airsim}
S.~Shah, D.~Dey, C.~Lovett, and A.~Kapoor, ``Airsim: High-fidelity visual and physical simulation for autonomous vehicles,'' in \emph{Field and Service Robotics: Results of the 11th International Conference}.\hskip 1em plus 0.5em minus 0.4em\relax Springer, 2018, pp. 621--635.

\bibitem{ArtStation}
\BIBentryALTinterwordspacing
K.~Sibiriakov, ``Artstation page https://www.artstation.com/vegaart,'' 2022. [Online]. Available: \url{https://www.artstation.com/artwork/ZGrnzN}
\BIBentrySTDinterwordspacing

\bibitem{UrbanScene3D}
L.~Lin, Y.~Liu, Y.~Hu, X.~Yan, K.~Xie, and H.~Huang, ``Capturing, reconstructing, and simulating: the urbanscene3d dataset,'' in \emph{ECCV}, 2022.

\bibitem{sitzmann2019siren}
V.~Sitzmann, J.~N. Martel, A.~W. Bergman, D.~B. Lindell, and G.~Wetzstein, ``Implicit neural representations with periodic activation functions,'' in \emph{arXiv}, 2020.

\bibitem{1284395}
Z.~Wang, A.~Bovik, H.~Sheikh, and E.~Simoncelli, ``Image quality assessment: from error visibility to structural similarity,'' \emph{IEEE Transactions on Image Processing}, vol.~13, no.~4, pp. 600--612, 2004.

\bibitem{zhang_unreasonable_2018}
R.~Zhang, P.~Isola, A.~A. Efros, E.~Shechtman, and O.~Wang, ``The {Unreasonable} {Effectiveness} of {Deep} {Features} as a {Perceptual} {Metric},'' in \emph{2018 {IEEE}/{CVF} {Conference} on {Computer} {Vision} and {Pattern} {Recognition}}, June 2018, pp. 586--595, iSSN: 2575-7075.

\bibitem{schonberger_structure--motion_2016}
J.~L. Schönberger and J.-M. Frahm, ``Structure-from-{Motion} {Revisited},'' in \emph{2016 {IEEE} {Conference} on {Computer} {Vision} and {Pattern} {Recognition} ({CVPR})}, June 2016, pp. 4104--4113.

\end{thebibliography}

\end{document}